# Brain Co-Processors: Using AI to Restore and Augment Brain Function

Rajesh P. N. Rao
Center for Neurotechnology &
Paul G. Allen School of Computer Science and Engineering
University of Washington, Seattle

## Abstract

Brain-computer interfaces (BCIs) use decoding algorithms to control prosthetic devices based on brain signals for restoration of lost function. Computer-brain interfaces (CBIs), on the other hand, use encoding algorithms to transform external sensory signals into neural stimulation patterns for restoring sensation or providing sensory feedback for closed-loop prosthetic control. In this article, we introduce *brain co-processors*, devices that combine decoding and encoding in a unified framework using artificial intelligence (AI) to supplement or augment brain function. Brain co-processors can be used for a range of applications, from inducing Hebbian plasticity for rehabilitation after brain injury to reanimating paralyzed limbs and enhancing memory. A key challenge is simultaneous multi-channel neural decoding and encoding for optimization of external behavioral or task-related goals. We describe a new framework for developing brain co-processors based on artificial neural networks, deep learning and reinforcement learning. These "neural co-processors" allow joint optimization of cost functions with the nervous system to achieve desired behaviors. By coupling artificial neural networks with their biological counterparts, neural co-processors offer a new way of restoring and augmenting the brain, as well as a new scientific tool for brain research. We conclude by discussing the potential applications and ethical implications of brain co-processors.

## Keywords

## Introduction

A brain-computer interface (BCI) [1, 2, 3, 4] (also called brain-machine interface or BMI) is typically defined as a device that "reads" (or "decodes") signals from the brain to directly control external devices such as prosthetics, cursors or robots. A computer-brain interface (CBI), on the other hand, can be defined as a device that "encodes" external signals to be delivered to the brain and "writes" that information to the brain through neural stimulation. The origins of BCIs and CBIs can be traced to neuroscience research in the 1960s and 1970s by Fetz [4], Delgado [6], and Vidal [7]. For example, Eberhard Fetz used a simple type of BCI to study operant conditioning in monkeys: he trained monkeys to control the movement of a needle in an analog meter by modulating the firing rate of a motor cortical neuron.

There was a resurgence of interest in the field starting in the 1990s sparked by the availability of multi-electrode arrays and fast, cheap computers [1, 2]. Researchers developed BCIs based on increasingly sophisticated machine learning techniques for decoding neural activity for controlling prosthetic arms [8,9,10], cursors [11, 12, 13, 14, 15, 16], spellers [17, 18] and robots [19, 20, 21, 22]. In parallel, CBIs were developed for encoding and delivering artificial sensory information via stimulation to the brain and other regions of the nervous system for auditory [23], visual [24], proprioceptive [25], and tactile [26, 27, 28, 29, 30] perception.

These advances in neural decoding and encoding have opened the door to a new class of brain interfaces which we call *brain co-processors*: these interfaces leverage artificial intelligence (AI) to combine decoding and encoding in a single system and optimize neural stimulation patterns to achieve specific goals. Brain interfaces that combine recording and stimulation have been called bi-directional (or recurrent) BCIs [1] but these interfaces have typically used separate BCI and CBI components, making them special cases of the more unified concept of brain co-processors that we explore in this chapter.

We first review research illustrating how the framework of brain co-processors can be used for closed-loop control of prosthetic devices, reanimation of paralyzed limbs, restoration of sensorimotor and cognitive function, neuro-rehabilitation, enhancement of memory, and brain augmentation, e.g., direct brain-to-brain interaction.

We then describe a new type of brain co-processors called "neural co-processors" [31] based on artificial neural networks, deep learning and reinforcement learning. Neural co-processors are designed to jointly optimize cost functions with the nervous system to achieve goals such as targeted rehabilitation and augmentation of brain function, and provide a new scientific tool for testing computational models and uncovering principles of brain function [32]. The chapter concludes with a discussion of potential applications and ethical implications of brain co-processors as researchers and commercial enterprises increasingly look to augmentative applications of such co-processors.

**Brain Co-Processors**

Figure 1 depicts the general form of a brain co-processor. A co-processor uses AI to transform neural activity <u>and/or</u> external inputs into stimulation patterns <u>and/or</u> external control signals for actuators.

Inputs to the co-processor may include:
- Neural recordings, e.g., spikes or local field potentials (LFPs) from microlectrodes, electrical activity from electrocorticography (ECoG) or electroencephalography (EEG), optical/optogenetic recordings, and blood flow changes from fMRI or fNIRS.
- External information sources such as sensors (infrared, ultrasonic etc.), the Internet, local storage device providing data, a source of user input, or even another nervous system.

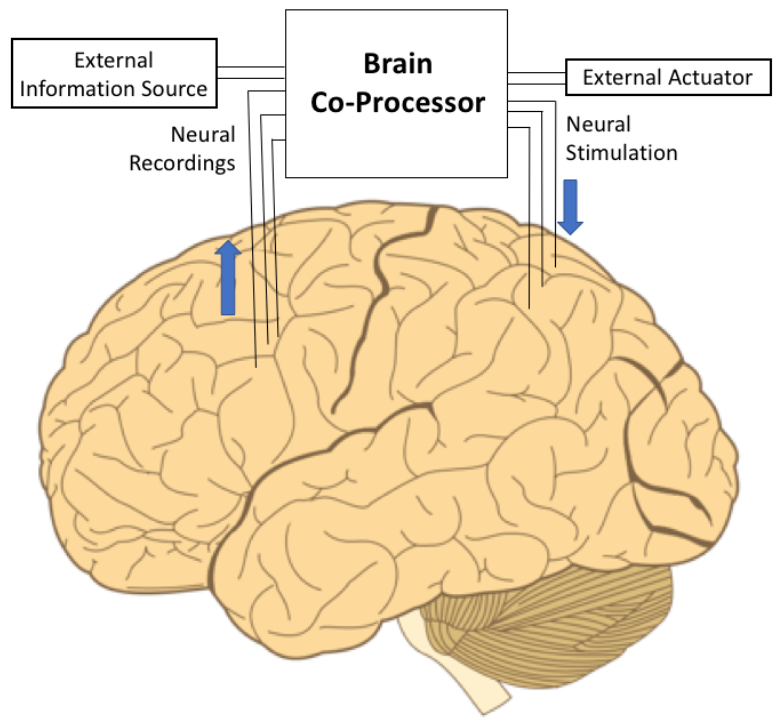

**Figure 1: Brain Co-Processor.** The figure depicts the general form of a brain co-processor. The co-processor receives as input both external information from sensors, the internet or other information source as well as ongoing neural activity. The output of the co-processor includes commands to external actuators (robots, internet search, message to another brain, etc.) as well as multi-dimensional stimulation patterns delivered to one or more regions of the nervous system to achieve a desired goal.

Outputs of the co-processor include:
- multi-dimensional neural stimulation patterns, e.g., stimulation delivered electrically via microelectrode or ECoG arrays, optical stimulation via optogenetic techniques, focused ultrasound stimulation, and magnetic stimulation.
- control signals for an external actuator such as commands for a robot, a computing device, an internet search engine, or messages to another brain.

The algorithms implemented on the co-processor for transforming its inputs into suitable outputs can be AI algorithms ranging from simple mappings or rules based on medical or other domain knowledge to sophisticated machine learning algorithms for classification, regression, or reinforcement learning. Specifically, the sophistication of the transformation may range from a simple identity mapping (each input spike results in a stimulation pulse) to complex mappings mediated by artificial neural networks with modifiable weights (see *Neural Co-Processors* below).

In the next section, we review examples of brain co-processors that have been proposed for (a) restoring lost function and (b) augmenting natural human function.

**Brain Co-Processors for Restoring Lost Function**
Brain co-processors can be used for restoring sensory or motor function, for example, restoring vision, controlling a robotic arm, reanimating a paralyzed limb, modulating neural circuits for alleviating the symptoms of motor or cognitive disorders, and inducing neuroplasticity for targeted rehabilitation of the injured brain. We review some of these examples of brain co-processors in this section.

*Closed-Loop Sensory Prostheses*
Figure 2A shows the traditional approach to a sensory prosthesis, such as a visual prosthesis for the blind. Images from a camera are transformed into stimulation patterns via a computer-brain interface (CBI) implementing a simple encoding algorithm, e.g., converting grayscale pixel values averaged over an image region to a proportional frequency or amplitude value for stimulation pulses that are delivered to a visual region in the brain such as the visual cortex [33]. The results from such devices have been mixed. One reason is likely the fact that this approach does not take into account the underlying dynamics of neural activity in the region being stimulated. As a result, the effects of the exact same stimulation parameters on a neural population may vary from one stimulation instance to another due to the ongoing neural dynamics, resulting in differences in perception. Similar limitations apply to stimulation-based approaches for artificial sensation in other modalities such somatosensory stimulation [34, 35].

The brain co-processor approach to designing sensory prostheses is depicted in Figure 2B. The co-processor receives as input not only the sensor values (e.g., image pixel values from a camera) but also the current neural recordings from the region being stimulated or areas near this region, as well as neural recordings from other areas pertinent to perception of the stimulus (e.g., higher-order sensory areas, motor regions involved in sending predictive sensory signals, etc.). Thus, by design, the co-processor's AI algorithm takes into account both the ongoing neural dynamics and the external sensory input to compute stimulation patterns appropriate for current brain state in order to achieve a reliable percept. The AI algorithm's parameters (e.g., weights of artificial neural networks) can be tuned based on the subject's feedback to optimize the parameters for reliable perception (see *Neural Co-Processor* section below).

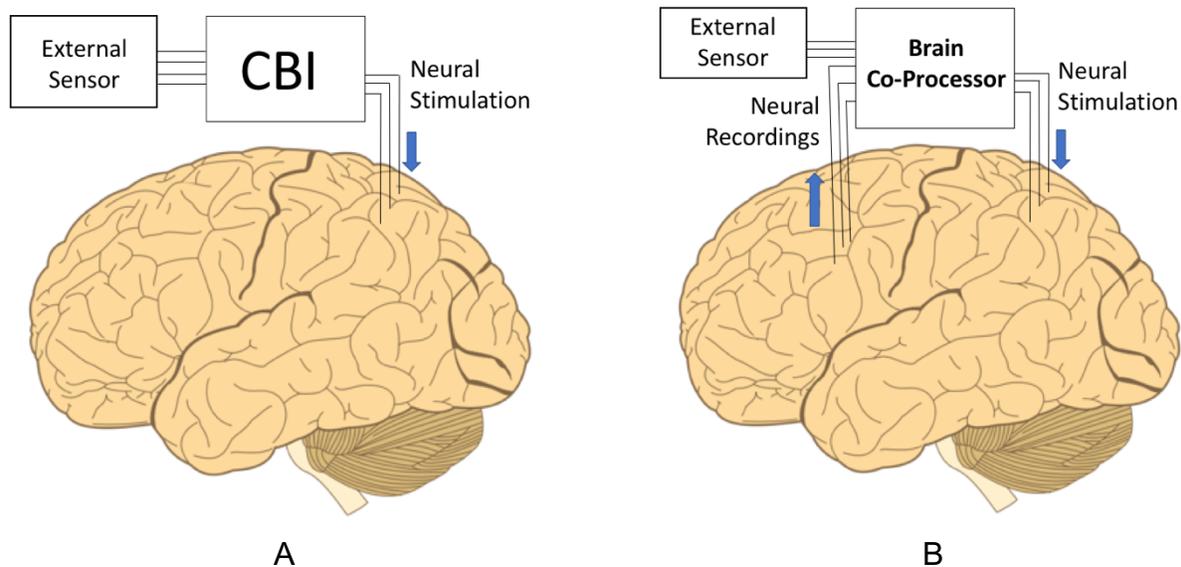

**Figure 2: Sensory Prostheses.** (A) Traditional sensory prosthesis, e.g., a visual prosthesis, based on a computer-brain interface (CBI). (B) Brain co-processor for sensory restoration/augmentation: Besides encoding sensor information, the co-processor takes into account ongoing dynamics of the brain to tailor its stimulation pattern for reliable perception.

*Closed-Loop Prosthetic Control*

Traditional BCIs (Figure 3A) decode neural recordings from one or more brain regions (e.g., motor cortex) to control prosthetic devices such as a robotic arm or hand for restoring motor function or a cursor for communication. The AI algorithm used for decoding is frequently the Kalman filter [36, 37]. Typically, the unknown kinematic quantities to be estimated, such as hand position, velocity, and acceleration, are represented by a "state vector" $\mathbf{x}$.

The likelihood (or measurement) model for the Kalman filter specifies how the "hidden" kinematic vector $\mathbf{x}_t$ at time $t$ relates (linearly via a matrix $B$) to the measured neural activity vector $\mathbf{y}_t$:

$$\mathbf{y}_t = B\mathbf{x}_t + \mathbf{m}_t$$

The dynamics model specifies how $\mathbf{x}_t$ linearly changes (via matrix $A$) over time:

$$\mathbf{x}_t = A\mathbf{x}_{t-1} + \mathbf{n}_t$$

$\mathbf{n}_t$ and $\mathbf{m}_t$ are zero-mean Gaussian noise processes. The Kalman filter computes for each time state $t$ an optimal estimate for the kinematics vector $\mathbf{x}_t$ (both mean and covariance) given the current measurement $x_t$ and all past measurements – the equations for computing the mean and covariance can be found in [37].

A Kalman filter can be used in a BCI such as the one depicted in Figure 3A for decoding brain activity for controlling a prosthetic device – the subject in this case relies on visual feedback to alter brain activity for closed-loop control. However, visual feedback alone is insufficient for precise control of a prosthetic, e.g., to determine the force to be applied to hold a paper cup versus as ceramic cup, or to detect slippage of an object being held. In such cases, the brain requires feedback from artificial tactile and proprioceptive sensors on the prosthetic device.

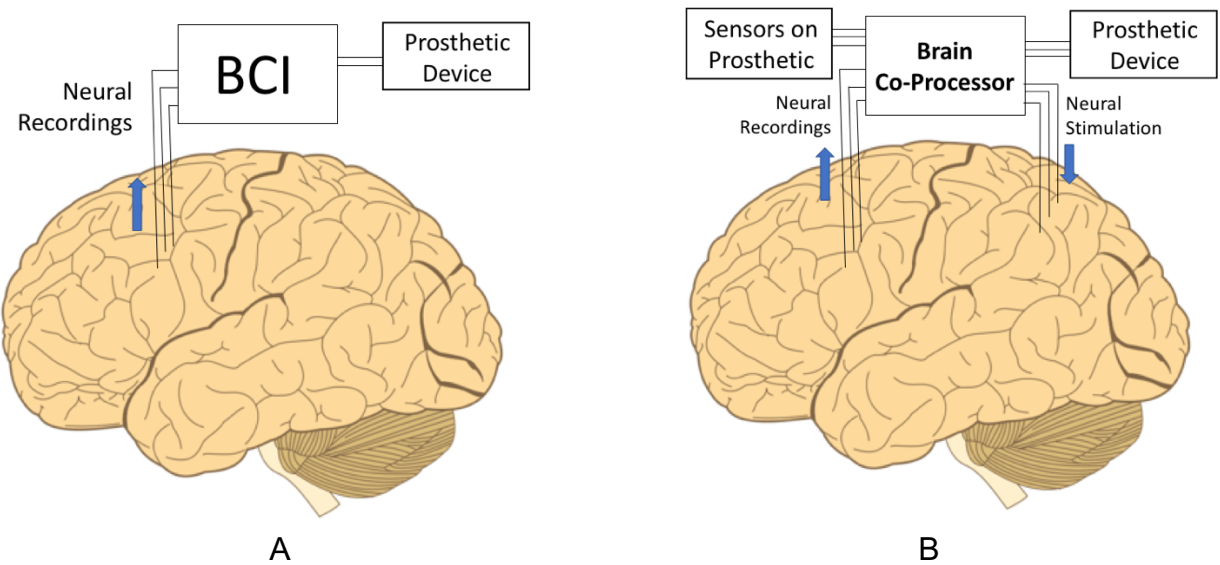

**Figure 3: Prosthetic Control.** (A) Traditional BCI or BMI (brain-machine interface) for prosthetic control. (B) Brain co-processor for closed-loop prosthetic control.

The brain co-processor shown in Figure 3B allows closed-loop control of a prosthetic device by combining recording and stimulation in an integrated framework. The inputs to the co-processor include external measurements from tactile and proprioceptive sensors, as well as neural signals from both motor and sensory regions of the brain. The motor neural signals are decoded by an AI algorithm such as the Kalman filter described above to generate control signals for the prosthetic hand. The same algorithm or a different method is used to encode information from the artificial sensors on the prosthetic device, in conjunction with neural recordings in sensory areas (see above), to appropriately stimulate somatosensory neurons for tactile and proprioceptive feedback. An important requirement for such a co-processor is that stimulation-generated artifacts should not corrupt the recorded signals being used to control the hand; we refer the reader to [38, 39, 40, 41] for examples of methods for handling such stimulation artifacts.

Although the co-processor framework of Figure 3B is yet to be fully tested, several research groups have explored versions of this co-processor framework for prosthetic control. For example, researchers have shown that stimulation of somatosensory cortex can be used to instruct a rhesus monkey which of two targets to move a cursor to; the cursor is subsequently controlled using linear decoding to predict the X- and Y-coordinate of the cursor [42].

In a subsequent study demonstrating closed-loop control [43], monkeys used a BCI based on primary motor cortex (M1) recordings and Kalman-filter-based decoding to actively explore virtual objects on a screen with artificial tactile properties. The monkeys were rewarded if they found the object with particular artificial tactile properties. During brain-controlled exploration of an object, the associated tactile information was delivered to somatosensory cortex (S1) via intracortical stimulation. Tactile information was encoded as a high-frequency biphasic pulse train (200 Hz for rewarded object, 400 Hz for others) presented in packets at a lower frequency (10 Hz for rewarded, 5 Hz for unrewarded objects). Because stimulation artifacts masked neural activity for 5–10 ms after each pulse, an interleaved scheme of alternating 50 ms recording and 50 ms stimulation was used. The monkeys were able to select the desired target object within a second or less based only on its tactile properties as conveyed through stimulation.

Monkeys can also utilize intracortical stimulation in S1 to perform a match-to-sample task where the goal is to move a virtual arm and find a target object that delivers stimulation similar to a control object [44]. In this experiment, the monkey controlled a virtual arm using a Kalman-filter-based decoding scheme where the Kalman filter's state was defined as the virtual hand's position, velocity and acceleration in three dimensions. The encoding algorithm involved stimulating S1 via three closely located electrodes using a 300 Hz biphasic pulse train for up to 1 second while the virtual hand held the object. After training, the monkey was able to move the virtual hand to the correct target with success rates between 70% and more than 90% over the course of 8 days (chance level was 50%).

In humans, closed-loop control of a prosthetic hand was achieved by a paralyzed subject in a continuous force matching task [45] using a co-processor-style framework. Control signals were decoded from multi-electrode recordings in M1 using a decoder based on a linear mapping of robotic arm movement velocities **v** to M1 firing rates **f**:

$$\mathbf{f} = B\mathbf{v}$$

where *B* denotes a matrix of weights estimated from data. Initial training data for the linear decoder was obtained by asking the subject to observe the robotic hand performing hand shaping tasks such as "pinch" (thumb/index/middle flexion-extension), "scoop" (ring/pinky flexion/extension) or grasp (all finger flexion) and recording M1 firing rates **f**, followed by a second training phase involving computer-assisted control to fine tune the decoder weights. The subject then performed a 2D force matching task with the robotic hand using the trained decoder to pinch, scoop or grasp a foam object either gently or firmly while using stimulation of S1 to get feedback on the force applied.

The encoding algorithm linearly mapped torque sensor data from the robotic hand's finger motors to pulse train amplitude of those stimulating electrodes that previously elicited percepts on the corresponding fingers of the subject. The subject was able to continuously control the flexion/extension of the pinch and scoop dimensions while evaluating the applied torque based on force feedback from S1 stimulation. The success rate for pinch, scoop or grasp with gentle or firm forces) was significantly higher with stimulation feedback compared to feedback from vision alone.

In all of these examples of closed-loop prosthetic control using BCIs, separate decoding and encoding methods were used; as mentioned in the previous section, encoding information for stimulation without taking into account the ongoing neural dynamics may lead to unreliable feedback. We address this issue below using brain co-processors called *neural co-processors* that use artificial neural networks to combine decoding and encoding within a single AI framework.

*Reanimating Paralyzed Limbs*
A brain co-processor can be used for reanimating a paralyzed limb by translating motor commands from the brain to stimulation patterns for spinal neurons (Figure 4) or muscles. As an example, a co-processor for brain-based control of electrical stimulation of muscles was used to restore movement in two monkeys [46]. The activity of single motor cortical neurons was converted to electrical stimulation of wrist muscles to move a cursor on a computer screen. The decoding scheme involved operant conditioning to train the monkey to volitionally control the activity of a selected motor cortical neuron to move a cursor into a target.

After training, the volitionally-controlled activity of the motor cortical neuron was converted into electrical stimuli which were delivered to the monkey's temporarily paralyzed wrist muscles – this type of stimulation is called functional electrical stimulation, or FES. Flexor FES current *Fl* was set to:

$Fl$ = 0.8 x [firing rate – 24] with a maximum of 10 mA.

Extensor FES was inversely proportional to the rate below a threshold:

$Ex$ = 0.6 x [12 – firing rate] with a maximum of 10 mA.

Both monkeys were able to modulate the activity of cortical neurons to control their paralyzed wrist muscles and move a manipulandum to acquire five targets. Ethier *et al.*

[47] extended these results to grasping and moving objects using a linear decoder with a static nonlinearity applied to about 100 neural signals from M1.

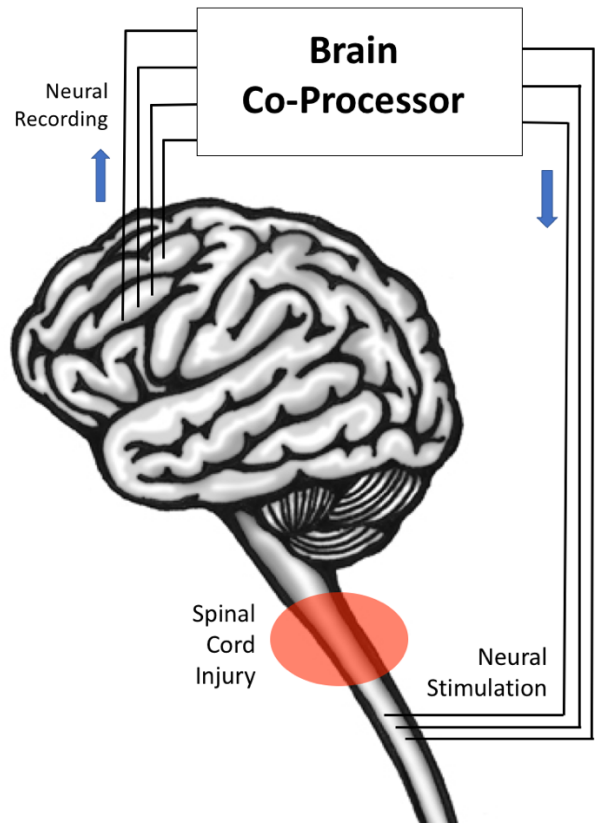

**Figure 4: Brain Co-Processor for Reanimation of Paralyzed Limbs.** Motor commands from regions of the brain such as the motor cortex can be processed by a brain co-processor and translated by the co-processor's AI algorithm to stimulation patterns, which are delivered to intact neural circuits in the spinal cord below the area of injury to reanimate a paralyzed limb.

In humans, Bouton *et al.* [48] showed that a quadraplegic man with a 96-electrode array implanted in the hand area of the motor cortex could use cortical signals to electrically stimulate muscles in his paralyzed forearm and produce six different wrist and hand motions. For decoding, six separate support vector machines (SVMs) were applied to mean wavelet power features extracted from multiunit activity to select one out of these six motions. The encoding scheme involved activating the movement associated with the highest decoder output using an electrode stimulation pattern previously calibrated to evoke that movement. Surface electrical stimulation was delivered as monophasic rectangular pulses at 50 Hz pulse rate and 500 μs pulse width, with stimulation intensity set to a piecewise linear function of decoder output.

These results were extended to multi-joint reaching and grasping movements by Ajiboye *et al.* [49]: a linear decoder similar to the Kalman filter described above was used to map neuronal firing rates and high frequency power at electrodes in the hand area of the motor cortex to percent activation of stimulation patterns associated with elbow, wrist or hand movements. The researchers showed that a tetraplegic subject could perform multi-joint arm movements for point-to-point target acquisitions with 80–100% accuracy and volitionally reach and drink a mug of coffee.

A shortcoming of the above approaches is that continued electrical stimulation of muscles results in muscle fatigue, rendering the technique impractical for day-long use. The brain co-processor approach in Figure 4 avoids this problem by using brain signals to stimulate the spinal cord rather than muscles. Spinal stimulation may additionally simplify encoding and control because it activates functional synergies, reflex circuits, and endogenous pattern generators. Progress in this direction was made by Capogrosso, Courtine, and colleagues [50] who demonstrated the efficacy of brain-controlled spinal stimulation for hind limb reanimation for locomotion in paralyzed monkeys. They used a decoder based on linear discriminant analysis to predict foot-strike and foot-off events during locomotion. The encoder used this prediction to activate extensor and flexor "hotspots" in the lumbar spinal cord via epidural electrical stimulation to correctly produce the extension and flexion of the impaired leg.

*Neuromodulation for Restoring Motor and Cognitive Function*
Figure 5 shows how a brain co-processor can be used to restore or replace lost function by modulating ongoing dynamics of neural circuits or by conveying information from one brain region to another bypassing an injured region. One of the early pioneers in this area was Jose Delgado [6] who designed an implantable BCI called the stimoceiver that could communicate with a computer via radio.

Delgado was the first to combine decoding with encoding to shape behavior: his decoding algorithm detected spindles in the amygdala of a monkey and for each detection, triggered stimulation in the reticular formation, which is associated with negative reinforcement. After six days, spindle activity was reduced to 1 percent of normal levels, making the monkey quiet and withdrawn. Unfortunately, efforts to extend this approach to humans to treat depression and other disorders yielded inconsistent results, possibly due to the lack of availability of large-scale and precise recording/stimulation techniques, computational processing power, and sophisticated AI frameworks for decoding/encoding.

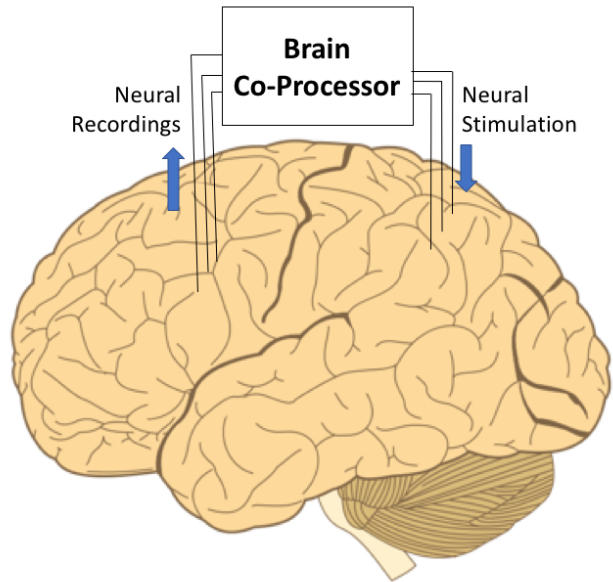

**Figure 5: Brain Co-Processor for Neuromodulation and Plasticity Induction.** A brain co-processor can translate neural recordings from one region of the brain to appropriate stimulation patterns delivered to another region of the brain for (a) modulating ongoing dynamics of neural circuits to correct undesirable behaviors and symptoms such as tremors, (b) replace lost function by emulating an injured neural circuit and conveying information from one brain region to another bypassing the injured region, and (c) induce neuroplasticity using the principle of Hebbian plasticity (see text for details).

Delgado's work did eventually inspire commercial brain implants such as Neuropace's RNS system that detects onset of seizures using time- and frequency-based methods from brain surface recordings (ECoG) and stimulates the region where the seizure originates. Also inspired by Delgado's work is the technique of deep brain stimulation (DBS), a widely prescribed form of neurostimulation for reducing tremors and restoring motor function in Parkinson's patients. Current DBS systems are open-loop but researchers have recently demonstrated closed-loop DBS (e.g., [51]) by triggering DBS based on movement intention, decoded as reduction in ECoG power in the low frequency ("mu") band over motor cortex.

*Inducing Plasticity and Rewiring the Brain*
The co-processor in Figure 5 can also be used for induction of plasticity and rewiring neural connections. Hebb's principle for plasticity states that if a group of neurons A consistently fires before another group of neurons B, connections from group A to group B should be strengthened since this indicates a causal relationship from A to B. Such plasticity can be artificially induced in the motor cortex of freely behaving primates by triggering stimulation at a site Nstim a few milliseconds after a spike was recorded at a different site Nrec (Figure 6A) [52]. In this case, the "AI" algorithm is a simple 1-to-1 mapping from one input spike to one output stimulation pulse.

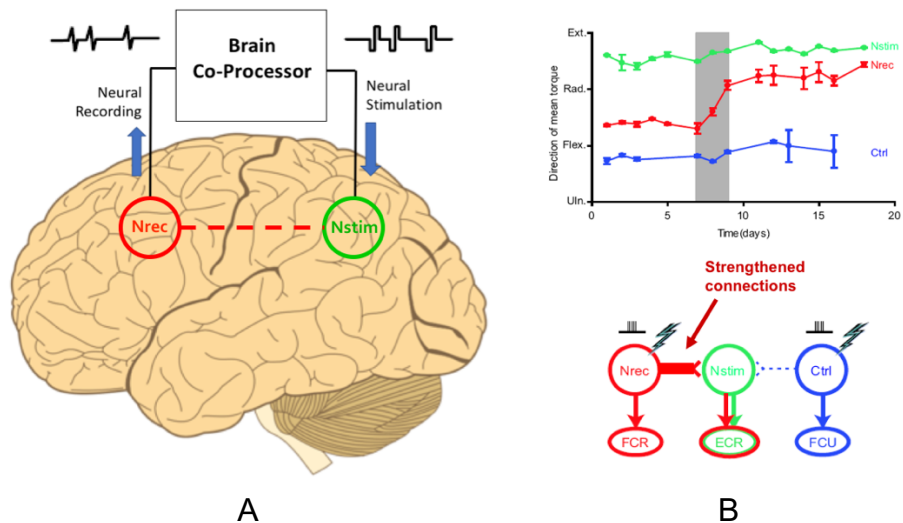

A                                           B

**Figure 6: Strengthening Connections between Cortical Neurons using a Co-Processor.** (A) A co-processor called the "Neurochip" delivered a biphasic stimulation pulse at the cortical site Nstim for each spike detected at the recording site Nrec. This artificial coupling of cortical sites continued during natural behavior for two days. The dashed line indicates possible weak existing connections between Nrec and Nstim. (B) After two days of the co-processor-mediated artificial connection from Nrec to Nstim (gray region, top panel), stimulation of Nrec was found to result in a mean torque (red) closer to the torque for Nstim (green), compared to stimulation of a control site (blue, Ctrl). The lower panel shows an explanation for these results. Adapted from [52].

After two days of such continuous spike-triggered stimulation, the output generated by site Nrec shifted to resemble the output from Nstim (Figure 6B, top panel), consistent with a strengthening of existing synaptic connections from neurons in Nrec to neurons in Nstim (Figure 6B, bottom panel).

The above method could be useful for neurorehabilitation since strengthening weak connections and more generally, rewiring the brain could allow restoration of brain function after traumatic brain injury, stroke or neuropsychiatric disorders such as depression and post-traumatic stress disorder (PTSD). As an example, Guggenmos, Nudo, and colleagues [53] used such an approach to improve reaching and grasping functions in a rat after traumatic brain injury to the rat's primary motor cortex (caudal forelimb area). An artificial connection was created by a co-processor bridging the rat's premotor cortex (rostral forelimb area or RFA) and somatosensory cortex S1. For each spike detected by an electrode in RFA, the co-processor delivered an electric pulse to S1 after 7.5 milliseconds, resulting in improvements in the rat's reaching and grasping performance.

The above approaches to plasticity induction relied on 1-to-1 spike-to-stimulation-pulse protocols. The question of how the approach can be generalized to induction of goal-directed multi-electrode plasticity is addressed with neural co-processors below.

**Brain Co-Processors for Augmenting Brain Function**
*Enhancing Memory*
Besides rehabilitation and restoration of lost function, brain co-processors can also be used for augmentation of existing brain function. As an example, a co-processor such as the one depicted in Figure 5 can be used to enhance short-term memory, as demonstrated by Berger, Deadwyler and colleagues [54, 55]. They implanted a co-processor system in the hippocampus of monkeys and rats to test memory enhancement in delayed match-to-sample (DMS) and nonmatch-to-sample tasks.

A multi-input multi-output (MIMO) nonlinear filtering model was first fit to simultaneously recorded spiking data from hippocampal CA3 and CA1 during successful trials, with CA3 activity as input and CA1 activity as output. Specifically, the spatiotemporal pattern transformations from CA3 to CA1 were learned by a MIMO model decomposed into a series of multi-input single-output (MISO) models with physiologically identifiable structure given by the equations:

$$w = u(k, x) + a(h, y) + \varepsilon(\sigma), \quad y = \begin{cases} 0 & \text{when } w < \theta \\ 1 & \text{when } w \geq \theta. \end{cases}$$

where:
- $w$ is a hidden variable that represents the pre-threshold membrane potential of the output model neurons, and is equal to the summation of three components, i.e. post-synaptic potential $u$ caused by input spike trains, the output spike-triggered after-potential $a$, and a Gaussian white noise $\varepsilon$ with standard deviation $\sigma$. The noise term models both intrinsic noise of the output neuron and the contribution of unobserved inputs. When $w$ exceeds threshold, $\theta$, an output spike is generated and a feedback after-potential ($a$) is triggered and then added to $w$.
- The variable $x$ represents input spike trains;
- $y$ represents output spike trains;
- Feedforward kernels $k$ describe the transformation from $x$ to $u$;
- The feedback kernel, $h$, describes the transformation from $y$ to $a$;
- $u$ can be expressed as a Volterra functional series of $x$ (see [54] for details), and $a$ is given by:

$$a(t) = \sum_{\tau=1}^{M_h} h(\tau) y(t - \tau),$$

The trained MIMO model was later applied to CA3 activity to predict CA1 activity converted to biphasic electrical pulses. In the four monkeys tested [55], performance in the DMS task was enhanced in the difficult trials, which had more distractor objects or required information to be held in memory for longer durations. One drawback of such an approach, which we address with neural co-processors below, is that we typically do not have the type of training data used to train the MIMO model because the brain is not healthy anymore by the time the device is needed (e.g., in Alzheimer's patients [56]), i.e., we do not have simultaneous recordings from areas such as CA3 and CA1 from the time that the brain was healthy.

*Brain-to-Brain Interfaces*
Figure 7 depicts how brain co-processors can be used to augment human communication and collaboration capabilities by facilitating direct brain-to-brain interactions.

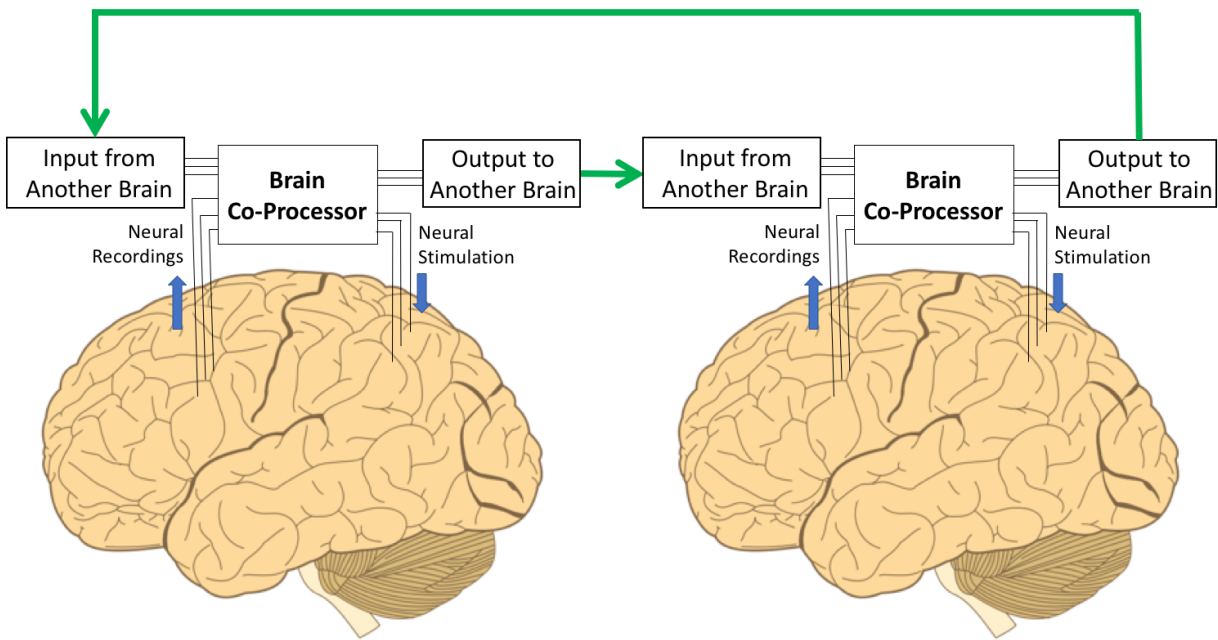

**Figure 7: Brain Co-Processors for Direct Brain-to-Brain Interaction.** Each person utilizes a co-processor to send information to one or more other brains and receive information from these brains. The co-processor is optimized to reliably interpret and encode the signals from another brain for stimulation and reliably decode information from one's own brain for transmission to another brain. See text for details and examples.

The first such human brain-to-brain interactions were demonstrated by Rao, Stocco and colleagues utilizing noninvasive recording and stimulation technologies to build brain-to-brain interfaces [57, 58, 59]. Electroencephalography (EEG) was used to decode from the motor or visual cortex the intention of a "Sender" who could perceive but not act; this decoded intention was delivered via transcranial magnetic stimulation (TMS) to the motor or visual cortex of a "Receiver" who could act but not perceive. The researchers showed that tasks such as a video game [57] or "20 questions" [59] could be completed successfully through direct brain-to-brain collaboration (see [60, 61] for other examples).

Figure 8 shows the performance of three pairs of humans using the first human brain-to-brain interface [57]. The Sender and Receiver played a video game which required a moving rocket about to hit a city to be destroyed [57] using only brain signals. The Sender used motor imagery (imagining hand movement) to control a cursor to indicate the decision to destroy the rocket. An EEG BCI detected decreases in the Sender's EEG power in the "mu" band (typically 8–12 Hz) due to motor imagery and conveyed this decision directly to the Receiver's motor cortex via TMS, causing the Receiver's hand to move and hit a touchpad to destroy the rocket. As shown in Figure 8, two Sender-Receiver pairs (1 and 3) successfully transmitted information from one brain to another to cooperate and solve the task (i.e., the area under the red ROC curve is larger than 0.5), while pair 2 could not solve the task due to poor discriminability of the Sender's EEG signals.

Brain-to-brain interfaces have also been demonstrated by Nicolelis and colleagues in rats [62, 63]. In their experiments, an "encoder" rat identified a stimulus and pressed one of two levers while its motor cortex activity was transmitted to the motor cortex of a "decoder" rat [63]. The stimulation pattern was based on a Z score computed from the difference in the number of spikes between the current trial and a template trial. If the decoder rat made the same choice as the encoder rat, both rats were rewarded for the successful transfer of information between their two brains.

More recently, these approaches have been extended to create a network of brains or "BrainNet" allowing groups of humans [64] or rats [65] to collaborate and solve tasks together. Figure 9 illustrates the performance of the first human BrainNet [64] that allowed three human brains to cooperate to solve a task. In this case, the task was a simplified Tetris game that required two Senders to use a BCI based on the Steady State Visually Evoked Potential (SSVEP) to send their decision about whether or not to rotate a Tetris block to the Receiver. The decisions of the two Senders were delivered via two separate TMS pulses to the visual cortex. The Receiver then had to process the visual stimulation, experienced as phosphenes, and decide based on the Sender's decisions whether or not to rotate the block.

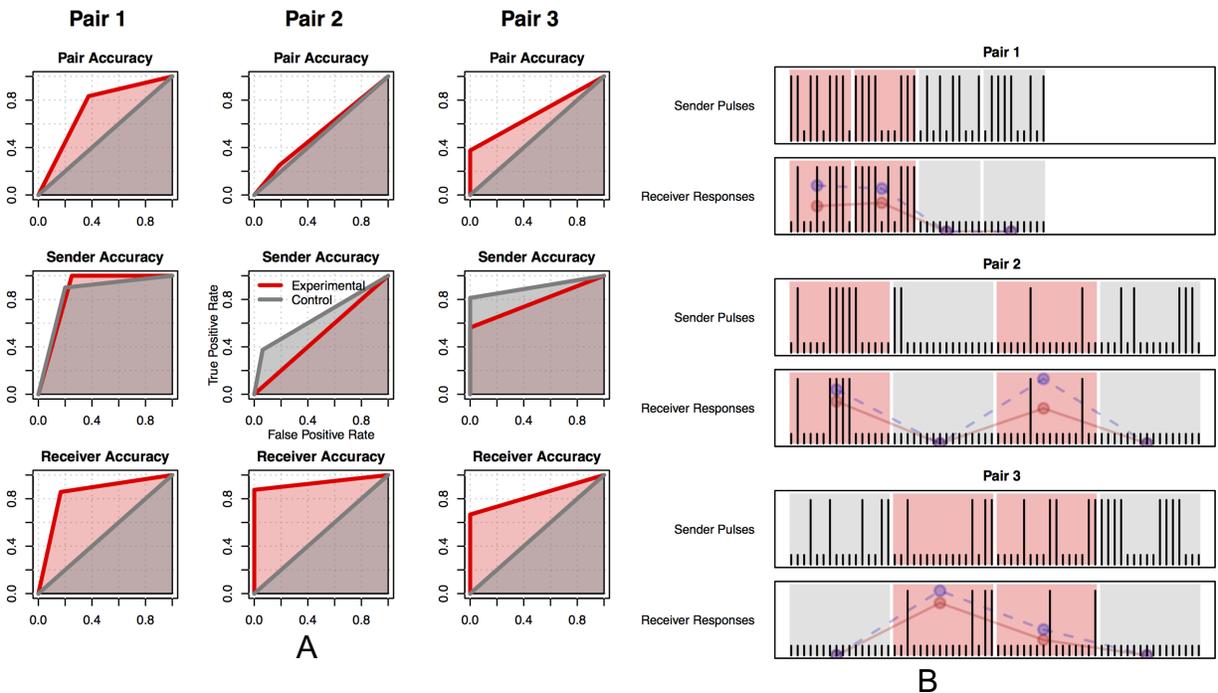

**Figure 8: Performance of the first Human Brain-to-Brain Interface.** (A) Receiver Operating Characteristic (ROC) curves (Y axis: true positive rate, X axis: false positive rate) for each of the three pairs of subjects (columns), presented in terms of overall pair accuracy (top panels), accuracy of the Sender (middle panels), and accuracy of the Receiver (bottom panels). Red lines and areas represent the "experimental" condition in which the brain-to-brain interface was operational, while grey lines and areas represent the control condition in which stimulation was inactivated. (B) Response Vectors of the Sender and Receiver. Each vertical tick represents a trial; long lines represent behavioral responses. Experimental blocks are marked by a red background; control blocks by a grey background. The blue dashed line represents the regression coefficient and the red line represents the mutual information between the Sender/Receiver Response vectors. Note that in all six experimental blocks (red background), the regression coefficient was significantly greater than zero while in all control blocks, the value was zero. About 4 to 13 bits of information were transferred from one brain to another during experimental blocks, compared to zero bits in the control blocks. (Adapted from [57]).

The Receiver could rotate the block using another EEG BCI based on SSVEP. The Senders could in turn perceive the rotation of the block (if any), and got one more chance to convey new decisions to the Receiver, allowing the triad of Senders-Receiver to correct any mistake made in the first round. Figure 9A shows the performance of the BrainNet in the form of ROC curves for five triads of subjects.

To test if the Receiver could learn from direct brain-to-brain interactions if one Sender was more reliable in their decisions than the other, the experimenters purposefully introduced errors in the decisions of one Sender, selected randomly as the designated "Bad Sender" across trials. Figure 9B shows that Receivers were indeed able to learn, over the course of several trials, which Sender was the reliable one and based their overall decision to rotate a block or not on the reliable Sender's decision, similar to how people in a social network form opinions about the reliability of people in the network. BrainNet can thus be regarded as a rudimentary form of a social network of brains.

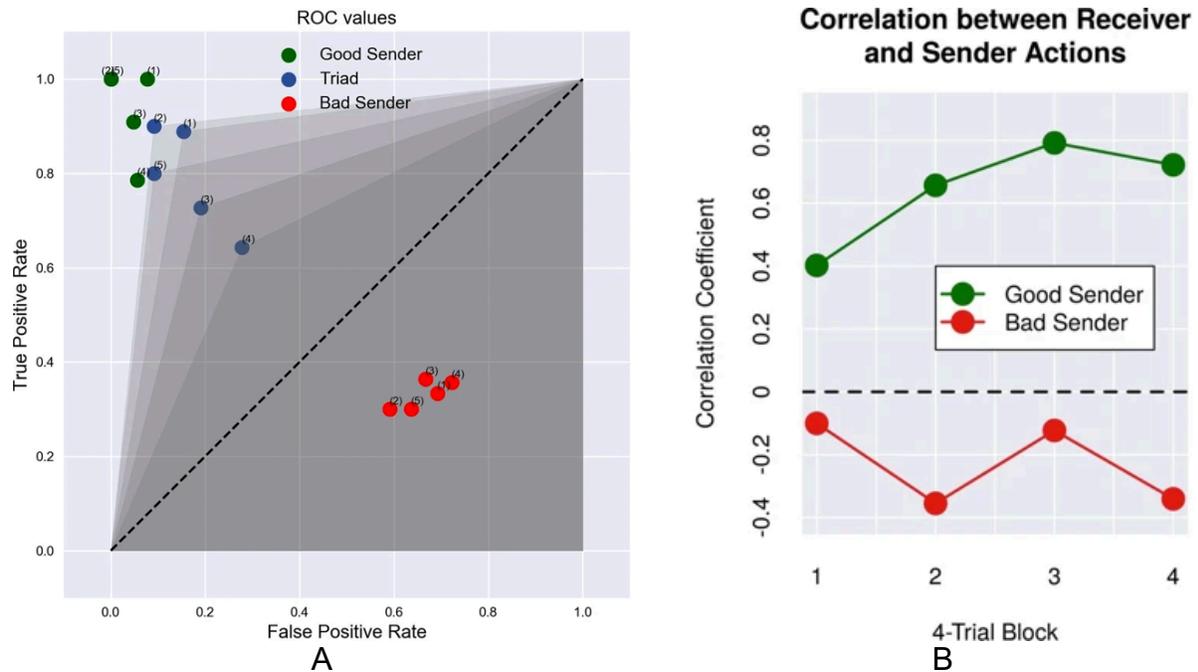

**Figure 9: Performance of the first Human BrainNet.** (A) ROC Curves for Five Triads of Participants (Two Senders and One Receiver in each Triad). The plot shows the overall performance of each triad (blue dots) as well as the performances of the two types of Senders ("Good" (reliable) versus "Bad" (unreliable)) in each triad (green and red dots). See [64] for the experimental design used to create a "Good" versus "Bad" Sender. The superscript on each dot denotes the triad number. Shaded areas represent the area under the curve (AUC) for each triad's ROC curve. The dashed line denotes chance performance. (B) Quantification of Learning of Sender's Reliability by Receiver. Evolution over time of Pearson Correlation Coefficient between the decisions of Receivers and Senders of each type. The plot exhibits ascending trends for the "Good" Sender but not the "Bad" Sender, suggesting that Receivers learned which Sender was more reliable based on their brain-to-brain interactions with the two Senders. (Adapted from [64]).

**Table 1. Summary of current applications of brain co-processors**

| Application | Input & Output | Algorithm | Limitations |
|---|---|---|---|
| Cursor control with artificial tactile feedback in monkeys [43] | Spikes from monkey M1 cortex & intracortical microstimulation in S1 cortex | Unscented Kalman filter & Biphasic pulse trains of different frequencies | Artificial tactile feedback limited to reward/no reward information |
| Brain-controlled prosthetic hand with force feedback in humans [45] | Spikes from human M1 cortex & intracortical microstimulation in S1 cortex | Velocity-based linear decoder & linear encoding of torque to pulse train amplitude | Simple force matching task with only two levels (gentle or firm) |
| Brain control of paralyzed muscles to restore wrist movement in monkeys [46] | Spikes from monkey M1 cortex & functional electrical stimulation of muscles | Volitional control of firing rate of single neuron & linear encoder | Simple flexion and extension movements of the wrist only, muscle fatigue with prolonged use |
| Brain control of forearm muscles for hand/wrist control in paralyzed human [48] | Multiunit activity from hand area of human M1 cortex & functional electrical stimulation of paralyzed forearm muscles | Support vector machines for classifying 1 of 6 wrist/hand motions & previously calibrated stimulation for each motion | Decoder based on classification of 6 fixed motions, muscle fatigue with prolonged use |
| Brain control of arm muscles for multi-joint movements & point-to-point target acquisitions in paralyzed human [49] | Spikes and high frequency power in hand area of human M1 cortex & functional electrical stimulation of paralyzed arm muscles | Linear decoder & percent activation of stimulation patterns associated with hand, wrist or elbow movements | Percent activation of fixed stimulation patterns, muscle fatigue with prolonged use |
| Brain control of the spinal cord for restoring locomotion in paralyzed monkeys [50] | Multiunit activity from leg area of monkey M1 cortex & epidural electrical stimulation of the lumbar spinal cord | Linear discriminant analysis to predict foot strike/foot off & activation of spinal hotspots for extension/flexion | Simple two state decoder and encoder models, viability for restoration of bipedal walking in humans yet to be demonstrated |
| First brain interface to control behavior and induce neuroplasticity in animals [6] | Local field potentials in monkey amygdala & electrical stimulation in the reticular formation | Decoder algorithm for detecting fast "spindle" waves & an electrical stimulation for each detection | Results not consistent from subject to subject, did not generalize to treating depression in humans |
| First demonstration of memory enhancement in a short-term memory task in a monkey [55] | Spikes from area CA3 in monkey hippocampus & electrical microstimulation in area CA1 in hippocampus | Multi-input/multi-output (MIMO) filtering model to decode CA3 activity & encode predicted CA1 activity as biphasic electrical pulses | Applicability to memory restoration in Alzheimer's or other patients unclear due to MIMO model training requirement |
| First demonstration of Hebbian plasticity induction in a freely behaving monkey [52] | Spikes from a region of monkey M1 cortex & intracortical microstimulation of a different region of M1 | Single spike detection & biphasic electrical pulse for each spike detection | Single input/single output protocol, not designed for multi-input/multi-output goal-directed plasticity induction |
| First demonstration of improved motor function after traumatic brain injury in a rat using plasticity induction [53] | Spikes from rat premotor cortex & intracortical microstimulation of S1 somatosensory cortex | Single spike detection in premotor cortex & electrical pulse stimulation in S1 after 7.5 ms. | Single input/single output protocol, plasticity induction not geared toward optimizing behavioral or rehabilitation metrics |
| First demonstration of brain-to-brain interfacing in humans [57] | Human EEG recordings in one subject and TMS stimulation over motor cortex in other subject | Changes in "mu" band (8–12 Hz) power due to hand motor imagery & TMS pulse to a pre-selected brain region that elicits upward jerk of hand positioned above a touchpad | Low information transmission rate of less than 1 bit/sec |

**Neural Co-Processors**

Most of the brain co-processor examples reviewed in the previous two sections treated decoding and encoding as separate processes, and did not co-adapt to jointly optimize a behavioral cost function with the nervous system.

We address these limitations by describing a type of brain co-processor called a *neural co-processor* (Figure 10) [31]. A neural co-processor uses two artificial neural networks, a co-processor network (CPN) and an emulator network (EN). A neural co-processor can be trained using either (i) a new type of deep learning [66] algorithm that approximates *backpropagation through both biological and artificial networks*, or (ii) a reward/cost-based *reinforcement learning* [67] algorithm.

*Training a Neural Co-Processor using Backpropagation-based Deep Learning*
Consider the problem of restoring movement in a stroke or spinal cord injury (SCI) patient, e.g., enabling the hand to reach a target object (see Figure 10). The CPN in this case could be a multi-layered (possibly recurrent) neural network that maps neural activity patterns from a large number of electrodes in areas A1, A2, etc. (e.g., movement intention areas spared by the stroke or injury) to appropriate stimulation patterns in areas B1, B2, etc. (e.g., intact movement execution areas in the cortex or spinal cord). When the subject forms the intention to move the hand to a target (e.g., during a rehabilitation session), the CPN maps the resulting neural activity pattern to an appropriate stimulation pattern.

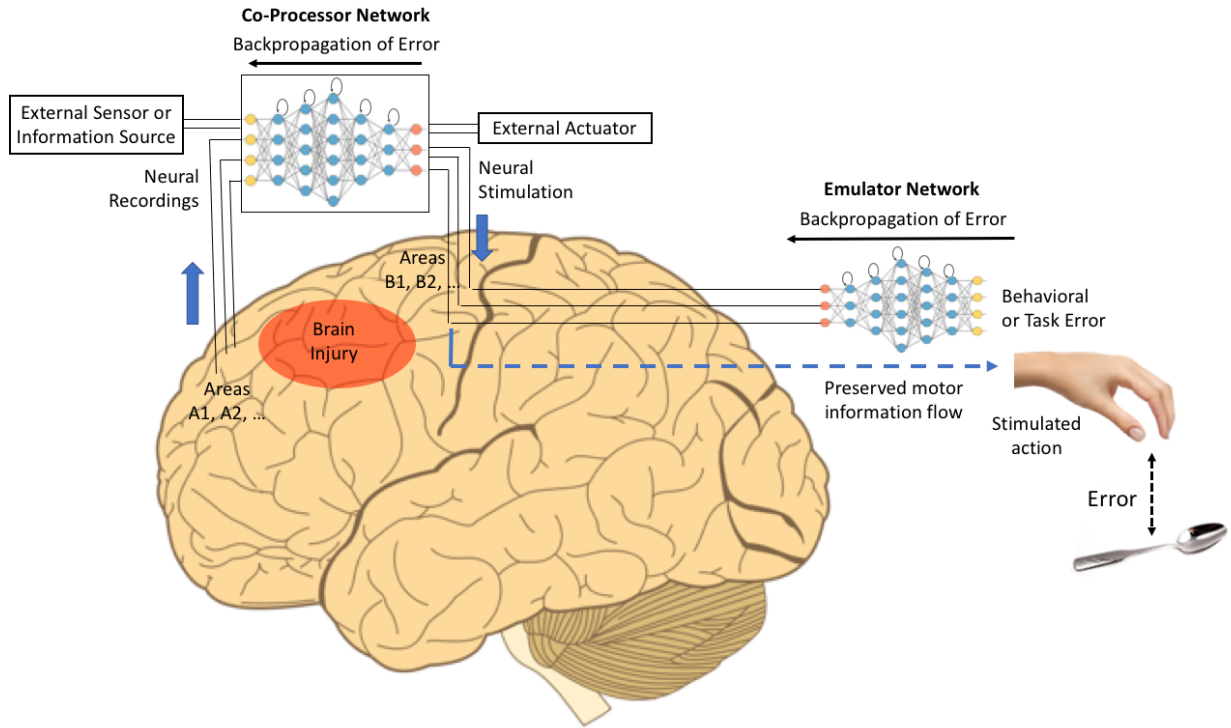

**Figure 10: Neural Co-Processor for Restoring and Augmenting Brain Function.** An artificial neural network called the "Co-Processor Network" (CPN) is used to map input neural activity patterns in one set of areas A1, A2, … to output stimulation patterns in other areas B1, B2, .... The CPN's weights are optimized to minimize brain-activity-based error (between stimulation patterns and target neural activity patterns *when known*), or more generally, to minimize behavioral/task error or maximize reward using another artificial neural network, an emulator network (EN). The EN is designed or pre-trained (e.g., via backpropagation) to learn the biological transformation from stimulation or neural activity patterns at the stimulation site to the resulting output behaviors. Using a trained EN, the CPN is trained to produce optimal stimulation patterns, thereby creating a goal-directed artificial information processing pathway between the input areas A1, A2, … and output areas B1, B2, .... External information from artificial sensors or other information sources can be integrated into the CPN's information processing as additional inputs to the neural network, and outputs of the CPN can include control outputs for external actuators. The example here shows the CPN creating a new information processing pathway between prefrontal cortex and motor cortex, bypassing an intermediate area affected by brain injury (e.g., stroke).

For simplicity, suppose the CPN is a three-layered network, with a single hidden layer (one layer only of blue circles in Figure 10). Let the matrix $V^{CPN}$ represent the weights from input layer of recorded neural activities $u^{CPN}$ to the hidden layer and let the matrix $W^{CPN}$ represent the weights from the hidden to the output layer representing stimulation patterns $v^{CPN}$. The output of the *i*th unit in the output layer can then be described as:

$$v_i^{CPN} = g(\sum_j W_{ji}^{CPN} g(\sum_k V_{kj}^{CPN} u_k^{CPN}))$$

where *g* is a nonlinear function such as the sigmoid function or the rectified linear unit (RELU) function [66]. If the desired stimulation patterns $d_i$ are known, we can find suitable weights $W^{CPN}$ and $V^{CPN}$ that minimize the error function:

$$E(W^{CPN}, V^{CPN}) = \frac{1}{2}\sum_i (d_i - v_i^{CPN})^2$$

This can be done using the "backpropagation" algorithm [66] which propagates the error $(d_i - v_i^{CPN})$ from the output layer back towards the input layer, updating first the weights $W^{CPN}$ and then the weights $V^{CPN}$ so as to minimize the error function *E*.

Unfortunately, in most scenarios, we do not know the correct stimulation patterns $d_i$ that produce a desired behavior, such as an intended hand movement. However, if the user pre-identifies a target such as a target location to move the hand to (e.g., during a rehabilitation session), then we can compute the error between the resulting hand movement after stimulation and the pre-identified target location, e.g., by using an overhead camera to determine the distance from the hand to the target. How can this behavioral error be used for backpropagation learning in the CPN to generate better stimulation patterns?

We address this problem using an emulator network (EN) that emulates the biological transformation from stimulation patterns to behavioral output. The EN is a deep (possibly recurrent) neural network whose weights can be learned using standard backpropagation. As an example, suppose the EN is a feedforward network of three layers, similar to the CPN above. Then, the output of the EN can be written as:

$$v_i^{EN} = g(\sum_j W_{ji}^{EN} g(\sum_k V_{kj}^{EN} u_k^{EN}))$$

where *W$^{EN}$* and *V$^{EN}$* are weights of the three-layered network (similar to the CPN case), and *u$^{EN}$* and *v$^{EN}$* denote the input and output of the network respectively, with the superscript EN denoting the fact that these parameters are for the emulation network.

Such an EN can be trained using a dataset consisting of a large variety of input stimulation (or neural activity) patterns *u$^{EN}$* applied to (or recorded in) areas B1, B2, etc. and the resulting movements or behavior $b_i$ measured using a sensor. We can then find, using backpropagation, the weights *W$^{EN}$* and *V$^{EN}$* that minimize the error function:

$$E(W^{EN}, V^{EN}) = \frac{1}{2}\sum_i (b_i - v_i^{EN})^2$$

After training, the EN acts as a surrogate for the biological networks mediating the transformation between inputs in B1, B2, etc. and output behavior.

An EN trained as described above can be used for learning the weights of the CPN in order to produce optimal stimulation patterns for minimizing behavioral error (e.g., error between current hand position and a target position). For each neural input pattern *X* (e.g., movement intention) that the subject produces in areas A1, A2 etc., the CPN produces an output stimulation pattern *Y* in areas B1, B2 etc., which results in a behavior *Z*. Since the EN has been optimized to make its output match the subject's behavior, the output *v*<sup>EN</sup> will approximate Z (any remaining error between *v*<sup>EN</sup> and Z can be further minimized offline by training the EN). We can therefore minimize the behavioral or task error between *Z* and the intended target behavior *Z*<sup>target</sup> by minimizing the following error function for the CPN:

$$E(W^{CPN}, V^{CPN}) = \frac{1}{2}\sum_i (Z_i^{target} - v_i^{EN})^2$$

This error function can be minimized using the backpropagation algorithm [66] by propagating the error $(Z_i^{target} - v_i^{EN})$ from the output layer of the EN back towards its input layer *but without modifying the EN's weights*. Once the backpropagated error reaches the input layer of the EN (which is the output layer of the CPN), we continue to backpropagate this error through the layers of the CPN, *but now modifying the CPN's weights for each of its layers*. In other words, the behavioral error $(Z_i^{target} - v_i^{EN})$ is backpropagated through a concatenated CPN-EN network but only the CPN's weights are changed. This allows the CPN to progressively generate better stimulation patterns that enable the brain to better achieve the target behavior with the CPN in the loop.

*Special Case: Using Only a Co-Processor Network*
A special case of the full neural co-processor framework described above is using only the CPN. A proof-of-concept of this special case was recently studied in [68]. This case is applicable when the target stimulation patterns $d_i$ are known and the EN is therefore not required. For example, if the goal is to suppress activation in a targeted region, e.g., to suppress a seizure, one can fix the target stimulation pattern to be a single fixed pattern or a small set of known stimulation patterns, and train the CPN to recognize specific patterns of input neural activity and map those activity patterns to the pre-determined target stimulation patterns. Similarly, if one has a good understanding of how stimulation in the target region affects behavior and other responses from the subject, e.g, via good computational models, one can use this knowledge to craft appropriate target stimulation patterns and use these stimulation patterns as the target $d_i$ to directly train the CPN without the need for an EN. Finally, if input and output neural activity patterns have been pre-recorded, as in the case of the MIMO approach described above [54], one can use this pre-recorded data to train the CPN to minimize the output stimulation error without requiring an EN. However, as mentioned above, most real-world scenarios will likely necessitate the use of an EN to train the CPN due to lack of knowledge of appropriate target stimulation patterns for achieving desired behaviors.

*Training using Reinforcement Learning and Rewards*
An alternate approach, in the absence of an error function, is to train the CPN using a form of reinforcement learning [67]. The CPN learns an optimal "policy" that maps input neural activity patterns in one set of regions (and possibly external inputs) to output stimulation patterns in other regions (and possibly outputs for actuators), as shown in Figure 10. The policy is optimal in terms of optimizing reward and/or cost, e.g., maximizing the total expected future reward as defined by an artificially-engineered reward function or by measuring natural reward signals such as dopamine in the brain. Reinforcement learning in this case solves the temporal credit assignment problem, i.e., favoring actions (stimulation patterns) that lead to desirable neural states in the future.

The reward/cost function can be designed by the co-processor designer to assign higher rewards to desirable neural activity patterns or desirable behavioral/task states. Undesirable states can be assigned penalties or high costs. If actual reward-related neural signals in the brain can also be recorded, such neural reward signals can be directly incorporated in the reinforcement learning reward function to optimize brain-based reward.

Given a reward/cost function, model-free or model-based reinforcement learning algorithms (e.g., Q-learning, temporal difference (TD) learning or actor-critic learning [67]) can be used to train the CPN to learn a policy that maximizes total expected future reward. To avoid the large number of trail-and-error trials typically required by model-free reinforcement learning (here requiring large amounts of stimulation of neural tissue), we can use model-based reinforcement learning based on an emulator network (EN) (as shown in Figure 10). The EN is designed from prior knowledge or trained to predict neural activity, rewards or behavior elicited by different stimulation patterns. The EN can be used as an approximate model or simulator to train the CPN using model-based reinforcement learning [67], thereby avoiding extensive stimulation of neural tissue during training.

*Plasticity Induction*
By repeatedly pairing patterns of neural inputs with patterns of output stimulation, the CPN can be expected to promote neuroplasticity between connected brain regions via Hebbian plasticity. Unlike previous plasticity induction methods [52, 53], the plasticity induced spans multiple electrodes and is goal-directed since the CPN is trained to minimize behavioral errors. After two regions A and B are artificially coupled for a sufficient amount of time, the resulting Hebbian plasticity may strengthen existing connections from A to B, causing neurons in region A to automatically recruit neurons in region B to achieve a desired response (such as a particular hand movement). Thus, in some cases, the neural co-processor may eventually no longer be required after a period of use and may be removed once function is restored or augmented to a satisfactory level.

*Beyond Motor Restoration: Cognitive and Sensory Restoration/Augmentation*
To illustrate the generality of the neural co-processor framework beyond restoring motor function, consider a co-processor for emotional well-being (e.g., to combat trauma, depression, or stress). The emulator network could first be trained by stimulating one or more emotion-regulating areas of the brain and noting the effect of stimulation on the subject's emotional state, as captured, for example, by a mood score based on a questionnaire answered by the subject [69]. The CPN could then be trained to map emotional intentions or other brain states to appropriate stimulation patterns that lead to a desired emotional state (e.g., less traumatic, stressful or depressed state).

Another example is using a neural co-processor to implement the sensory prosthesis depicted in Figure 2B, converting inputs from sensors such as a camera, microphone, or even infrared or ultrasonic sensors into stimulation patterns. In this case, the CPN in Figure 10 takes as input both external sensor information and current neural activity to generate an appropriate stimulation pattern in the context of the current state of the brain. The AI algorithm here could be a deep recurrent neural network, such as a long short-term memory (LSTM) network, that retains a memory of past brain states and sensor inputs, and combines this memory with the current sensor input to compute the optimal stimulation pattern. The emulator network in this case could be trained based on the subject's reports of perceptual states generated by a range of stimulation patterns.

More generally, the external input to the CPN could be from any information source, even the internet, allowing the brain to request information via the external actuator component in Figure 10. The resulting information is conveyed via the CPN's input channels and processed in the context of current brain activity using, for example, a recurrent neural network such as the LSTM network to determine the stimulation pattern. The emulator network in this case would be trained in a manner similar to the sensory prosthesis example above to allow the CPN to convert abstract information (such as text) to appropriate stimulation patterns that the user can understand after a period of learning, similar to how we learn to read based on visual stimulation patterns.

*Neural Co-Processors as Scientific Tools for Neuroscience*
Besides restoring or augmenting brain function, neural co-processors can be useful tools for testing new computational models of brain and nervous system function [32]. Rather than using traditional artificial neural networks in the CPN in Figure 10, one could use more realistic cortical models such as networks of integrate-and-fire or Hodgkin-Huxley neurons, along with biological learning rules such as spike-timing dependent plasticity rather than backpropagation. A critical test for putative computational models of the nervous system would then be: can the model successfully interact with its neurobiological counterpart and be eventually integrated within the nervous system's information processing loops? Similarly, one could replace the emulator network with a putative computational model of the neural systems governing the transformation from the stimulation site to the externally observed behavior. The accuracy with which the computational model mimics this input-output transformation can be evaluated based on the performance of the CPN, which relies on this computational emulation to generate appropriate stimulation patterns.

*Challenges*
A first challenge in realizing the above vision for neural co-processors is obtaining an error signal for training the two networks. In the simplest case (the special case discussed above), the error may simply be a neural error signal: the goal is drive neural activity in areas B1, B2, etc. towards known target neural activity patterns, and we therefore train the CPN directly to approximate these activity patterns without using an EN. However, we expect such scenarios to be rare.

In the more realistic case of restoring motor behavior, such as in stroke rehabilitation where the goal is, for example, to reach towards a known target pre-identified by a therapist, a computer vision system could be used to quantify the error between the hand position and the pre-identified target position. Similarly, a sensor worn on the hand could be used to indicate error in the force applied. Likewise, in speech rehabilitation after stroke, a speech analysis system could quantify the error between the generated speech and the target speech pattern. Finally, in the absence of an error signal, a reward/cost-based approach could be employed as described above in the section on *Training using Reinforcement Learning and Rewards*

An important question is how much training data is required to train the CPN and EN. If only the CPN is to be trained and the target stimulation patterns are known (see *Special Case* above), it may be possible to train the CPN based on pre-collected data, e.g., by recording neural activity for a period of time and labeling events of interest (e.g., seizures). The amount of training data will depend on the variability of the neural phenomena that the CPN needs to recognize, and a continual (possibly online) learning model may be required to account for co-adaptation of the brain.

In the more general case, it may be challenging to train an EN to be a sufficiently accurate model of the transformation from stimulation patterns to behavioral output. It may be difficult to obtain a sufficient amount of training data containing enough examples of how stimulation affects behavior since excessive stimulation may damage neural circuits. One possible solution is to record neural activities in regions that are causally related or correlated with observed behavior and use this data to train the EN, under the assumption that target stimulation patterns should approximate the recorded neural activity patterns. Another possible approach is to build the EN in a modular fashion, starting from biological structures closest to the target behavior and going up the hierarchy, e.g., learning to emulate the transformation from limb muscles to limb movements, spinal activity to muscle activity, etc. Finally, one could combine the above ideas with the concept of transfer learning using networks trained across similar neural regions or even across subjects [70], and incorporate prior knowledge from computational neuroscience models of the biological system being emulated (see the previous subsection). Regardless of the training method used, we expect that the EN (and CPN) will need to be regularly updated with new neural data as the brain adapts to having the CPN as part of its information processing loops.

**Applications and Ethical Implications of Brain Co-Processors**
*From In-Lab Demonstrations to Real-World Applications*
The co-processor experiments reviewed above mostly involved proof-of-concept demonstrations. While a small number of co-processors such as deep brain stimulators and Neuropace's RNS closed-loop stimulation system for controlling epilepsy are being used for medical applications, the vast majority of brain co-processors are still in their "laboratory testing" phase.

Consider, for example, the simplest variation of a brain co-processor, a BCI such as the one in Figure 3A that only records brain signals and decodes these signals to control a cursor. The maximum bit rate achieved by a human using such a system with invasive multi-neuron recordings is currently about 3.7 bits/sec and 39.2 correct characters per minute [16]. This is an order of magnitude lower than average human typing speeds of about 150-200 characters per minute. Considerable progress needs to be made in increasing the bandwidth and reliability of decoded information and making the risk-benefit ratio small enough in order to make real-world applications feasible. However, given the rapid strides in the field over the past decade and the entry of commercial entities in this space, co-processor applications may start appearing on the market within the next few decades, if not sooner. We discuss some of these potential applications below.

*Medical and Non-Medical Applications*
Figure 11 summarizes some of the medical and non-medical applications of brain co-processors. We have already discussed in previous sections some of the medical applications such as restoring motor, sensory or cognitive function via closed-loop control of prosthetic devices, closed-loop encoding for sensory prostheses, neuromodulation and plasticity induction for cognitive restoration and rehabilitation after stroke or other injury.

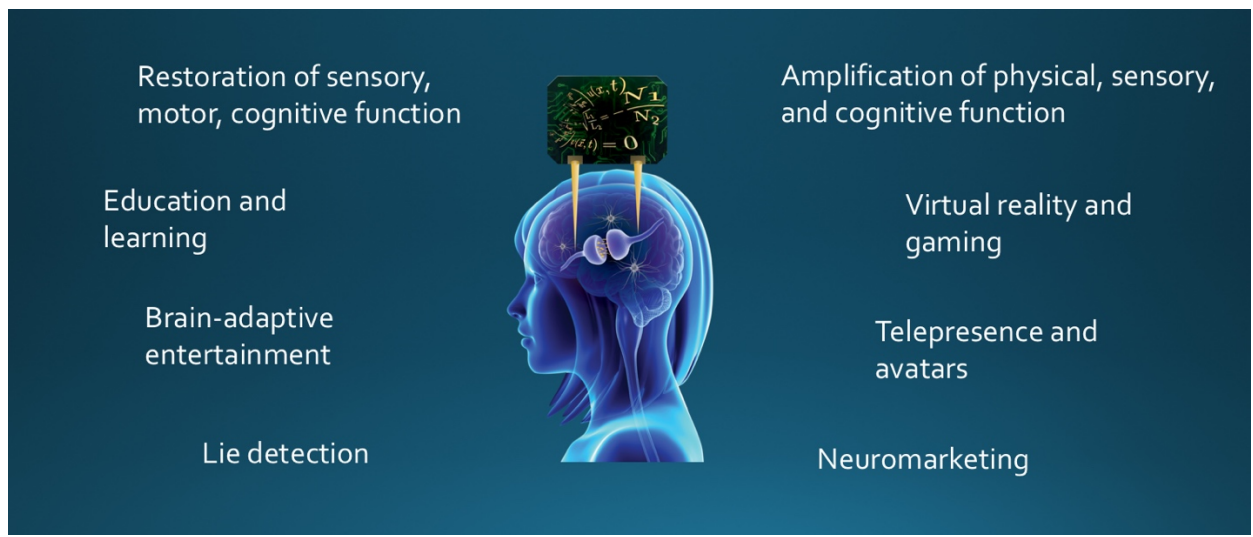

**Figure 11: Some Applications of Brain Co-Processors.**

Beyond medical applications, brain co-processors, such as the neural co-processor in Figure 10, could potentially be used in variety of ways to augment the capacity of the human brain:

- **Amplification of physical, sensory and cognitive function:** Co-processors could be used to amplify the physical capacity of a human through external actuators such as exoskeletons, robotic arms, or even "Ironman"-style body armor, with applications in firefighting, nuclear inspections, and maintaining law and order. Sensory amplification may include augmenting the brain with the ability to sense beyond the visible spectrum by providing as input to the brain co-processor measurements from infrared, hyperspectral, ultrasonic, laser-based, or other types of sensors. Cognitive amplification may be achieved in a variety of ways, e.g., by allowing the co-processor to augment the brain's knowledge and information processing capacity by rapidly accessing and integrating information from the internet.

- **Education and learning:** A brain co-processor could serve as an assistive device during knowledge acquisition by monitoring a student's progress, tracking attention, delivering lessons tailored to the student's optimal pace of learning, etc. Furthermore, the ability of co-processors to induce plasticity through closed-loop stimulation could potentially be used to accelerate the acquisition of knowledge and skills, or even transfer knowledge from an expert brain to another via brain-to-brain interfacing [58]. Eventually, the ability to verify whether a student has grasped the concepts in a course directly by monitoring corresponding changes in their brain activity may obviate the need for examinations and tests.

- **Virtual reality and gaming:** An obvious application is using brain co-processors for closed-loop brain stimulation for high fidelity virtual reality/augmented reality (VR/AR) and gaming. A proof-of-concept brain-stimulation-based VR game was described in [71]. Unlike today's VR/AR headsets that are limited to providing visual and auditory inputs, brain-stimulation-based VR systems would potentially allow a complete sensory experience including artificial smell, taste, proprioception, hunger, thirst, and somatic senses such as touch, heat, pressure and pain. Generating realistic sensations through stimulation will however require significant advances in our understanding of the neural basis of these sensations under natural circumstances.

- **Brain-adaptive entertainment:** Brain co-processors may open the door to personalized entertainment where the content may adapt not only to a person's overall preferences but also to current brain activity.

- **Telepresence and avatars:** The ability to touch, see and sense using sensor-rich robotic "avatars" in remote locations coupled to a co-processor conveying sensations through stimulation opens up the possibility of ultra-realistic telepresence, posing a potential threat to the air travel industry.

- **Lie detection and biometrics:** Methods for lie detection and "brain fingerprinting" for identification based on EEG and fMRI have already been proposed [72, 73], but the insufficient accuracy of these methods has prevented their adoption in law and policing. Closed-loop methods based on co-processors may eventually increase the accuracy of brain-based lie detection and biometrics to an acceptable level for real-world use, assuming the ethical issues can be satisfactorily addressed (see below).

- **Neuromarketing:** Marketing professional may be interested in gauging a person's response to an advertisement and in tracking a person's interest in a product by monitoring and studying their brain signals. A co-processor could potentially learn to track and even predict a person's interests over time as it interacts with the person's brain. While such an application is currently not feasible, it is nevertheless raises the important issue of ethics of these types of applications.

*Ethical, Moral and Social Justice Issues*
The possibility of augmenting human abilities with brain co-processors, as discussed in the previous section, brings to the forefront the urgent need to identify and address the ethical, moral and social justice issues before these technologies become feasible enough to be commercialized.

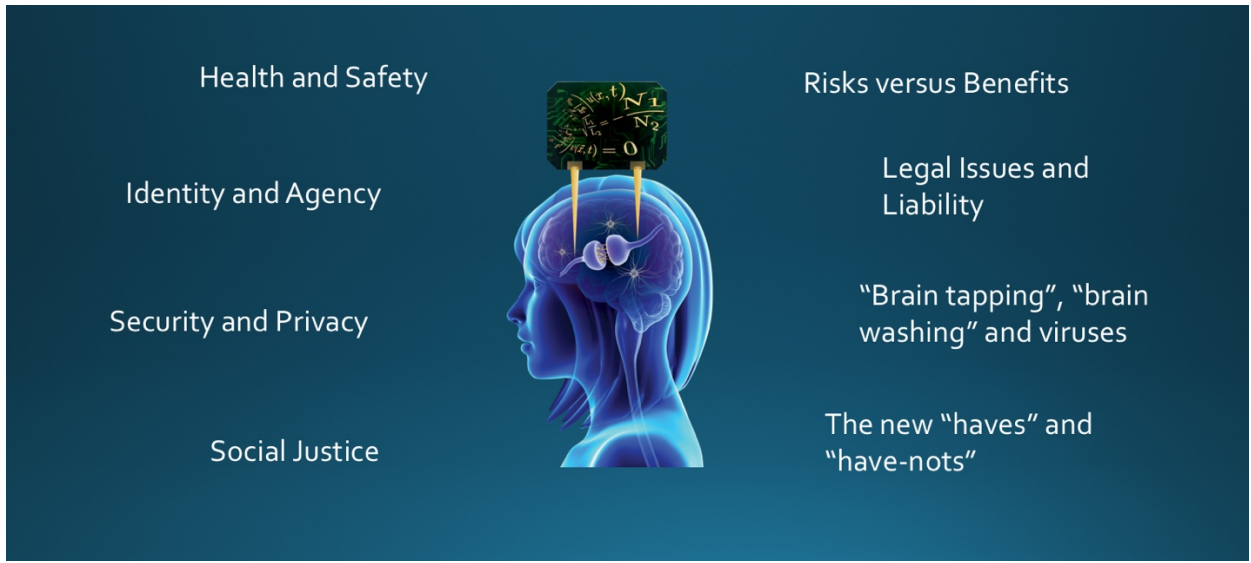

**Figure 12: Ethical, Moral and Social Justice Issues associated with the use of Brain Co-Processors.**

Figure 12 shows some of the major issues involved. These include:

- **Health and safety:** The most powerful co-processors will likely be invasive and require implantation, requiring the user to weigh the risks to health versus benefits of the technology. Does the increase in performance provided by an invasive co-processor (compared to a noninvasive one) justify, for a particular user, the increase in risk associated with invasive devices? Other questions to consider include the potential unintended side-effects of co-processor use, the potential for the user's expectations not being met after implantation, and the effects on family and caregivers.

- **Identity and agency:** The use of a co-processor has the potential to change a user's behavior in the long-term, thereby affecting their sense of identity [74]. Furthermore, in some cases, the user may feel that they have lost their sense of agency and ceded control to the co-processor [75]. Addressing and alleviating these potential threats to our notions of being human and having agency will be critical requirements for future co-processors.

- **Security and privacy:** Like most new technologies, there is a significant risk of brain co-processors being abused. Wireless communication from or to a brain could be intercepted ("brain tapping") and exploited by criminals, terrorists, commercial enterprises, or spy agencies as well as legal, law enforcement, and military entities. Brain stimulation opens up the dangerous possibility that an unsecure device may be hijacked and used to coerce a person to perform objectionable acts (e.g., commit a crime or sign a document such as a will). A device with access to memory-related regions in the brain could potentially be subverted to selectively erase memories [76] or write in false memories ("brainwashing"). Malicious entities could send a "virus" to a device, resulting in cognitive impairment or cognitive manipulation.

Given the potential for unprecedented abuse and malicious attacks, it is imperative that strong legal and technological safeguards are put in place before widespread deployment of any co-processor. Activities that violate a co-processor's security and privacy should be made illegal, with stringent punishments for breaking the law. Encryption techniques and security methods will need to have much stronger guarantees against attacks than current techniques and methods.

- **Legal issues:** Given the scope for abuse, lawmakers will need to pass sufficiently nuanced legislation to regulate what type of co-processors are legal to use and what are not – this may vary from country to country similar to how laws governing controlled substances today are different in different countries. Additionally, liability laws may need to change - courts will need to decide who is responsible for accidents or unlawful acts committed using a co-processor. Since co-processors use AI to adapt and learn, it may not be clear if the law was broken due to a volitional command issued by the human user or due to the action of the AI. One possible solution is to place the responsibility entirely on users by asking them to sign a waiver before using a co-processor, absolving the co-processor company from liability except for manufacturing defects. Alternately, courts could maintain panels of AI experts charged with investigating whether a company is at fault due to the behavior of the company's AI algorithm.

- **Moral and social justice issues:** The use of co-processors as an integral part of the human brain has the potential to fundamentally redefine what it means to be human. Will some humans forego the advantages of augmenting their physical and mental capabilities and choose to live a co-processor-free existence? This could divide human society into a new type of "haves" and "have-nots." Furthermore, the rich might have their children implanted at an early age to give them an edge in mental and/or physical capabilities, leaving the poor behind, with potentially drastic social consequences. One potential solution is for governments to subsidize certain basic types of co-processors for those who otherwise would not be able to afford them, similar to free public education and healthcare in certain countries. Another moral dilemma arises from parents having to decide whether or not to implant their child to augment the child's *future* mental and/or physical capabilities. Is it ethical for parents to decide what type of augmentation a child should have? Is it ethical for them to opt out of such augmentation, potentially leaving the child at a significant disadvantage in the future compared to augmented children? Should there be different schools for students with and without cognitive enhancement? These questions challenge our current conceptions of what it means to be human and point to the need for a comprehensive discussion of these issues among all stakeholders.

We hope that the ethical and moral issues raised above will help in the formulation of an internationally accepted code of regulations and ethics for co-processor development and future use.

**Conclusion**

Brain co-processors combine the concepts of brain-computer interfaces (BCIs) and computer-brain interfaces (CBIs) by using an AI-based framework to "close-the-loop" on the brain. The ability to simultaneously decode neural activity from one brain region and deliver information to another region via stimulation confers great versatility on co-processors. This chapter reviewed how co-processors can be used to (a) control prosthetic limbs with sensory feedback from artificial tactile sensors, (b) control paralyzed limbs through stimulation using motor intention signals from the brain, (c) restore and augment cognitive function and memory, and (d) induce neuroplasticity for rehabilitation. Promising proof-of-concept results have been obtained in animal models and in some cases, humans, but mostly under laboratory conditions. To transition to real-world conditions, co-processors must co-adapt with the nervous system and jointly optimize behavioral cost functions.

We introduced *neural co-processors* which use artificial neural networks to jointly optimize behavioral cost functions in synch with biological neural networks in the nervous system. A neural co-processor maps large multi-dimensional neural activity patterns (spikes, firing rates, local field potentials, ECoG, EEG etc.) to complex output stimulation patterns to achieve a desired goal. The simplest neural co-processor is a co-processor network (CPN) that is trained to output stimulation patterns that mimic a desired target neural activity pattern for each input. However, since a target stimulation pattern is typically unknown in realistic scenarios and only a behavioral or task-related goal may be known, one can use a second neural network called an emulator network (EN) and reinforcement learning or backpropagation learning to train the CPN. The trained EN acts as a surrogate for the biological network producing the neural or behavioral output, and allows the CPN to learn to deliver optimal stimulation patterns for specific input neural patterns.

Neural co-processors can be used to restore lost function. For example, in individuals with stroke or spinal-cord injury, a neural co-processor could be trained to map motor intention signals to uninjured regions of the brain or spinal cord to reanimate a paralyzed limb or restore function by inducing plasticity. Another application is mapping inputs from one memory-related area to another to facilitate or restore access to particular memories (e.g., in memory loss) or to unlearn traumatic memories (e.g., in PTSD). A related application is using a co-processor to unlearn unwanted behaviors (e.g., in obsessive compulsive disorders (OCD) or addiction) or retrain the brain in schizophrenia. Neural co-processors could also offer a more sophisticated alternative to current deep brain stimulation (DBS) approaches to treating depression and other neuropsychiatric disorders. Finally, the closed-loop co-adaptive capabilities of neural co-processors may offer a more powerful means to restore visual and other sensory function than current implants based on open-loop stimulation.

Besides medical applications, neural co-processors could potentially also be used for augmenting brain function. For example, one could map inputs from novel external sensors (e.g., infrared, ultrasonic etc.) to augment sensation and map brain signals to control signals for external actuators, closing the loop with the external world in a novel way. More generally, neural co-processors open up the possibility of augmenting the brain's knowledge, skills, information processing, and learning capabilities with the computational power of deep artificial neural networks for harnessing external information and guidance from sensors, the internet, and other brains.

Although co-processors are only now beginning to be validated in animal models and humans, it is imperative that we identify and address the ethical, moral, and social justice concerns associated with this new technology before it leaps too far ahead. In this chapter, we identified health and safety, identity and agency, security and privacy, and moral issues pertaining to the use of brain co-processors. The neural engineering community will need to work closely with ethicists, medical care providers, end users, policy makers, legal experts, and the general public in formulating appropriate guidelines and best practices for the development of safe, secure, ethically-informed and morally-grounded brain co-processors.

**Acknowledgments**
This work was supported by NSF grant EEC-1028725, CRCNS/NIMH grant no. 1R01MH112166-01, a CJ and Elizabeth Hwang endowed professorship and a grant from the W. M. Keck Foundation. The author would like to thank Eb Fetz, Chet Moritz, Andrea Stocco, Steve Perlmutter, Preston Jiang, Jon Mishler, and Richy Yun for discussions related to topics covered in this chapter.